\documentclass[11pt]{article}

\usepackage[preprint]{acl}

\usepackage{times}
\usepackage{latexsym}

\usepackage[T1]{fontenc}

\usepackage[utf8]{inputenc}

\usepackage{microtype}

\usepackage{inconsolata}

\usepackage{graphicx}

\usepackage{booktabs}
\usepackage{multirow}

\title{KDCM: Reducing Hallucination in LLMs through Explicit Reasoning Structures}

\author{Jinbo Hao\textsuperscript{1*}, Kai Yang\textsuperscript{2*}, Qingzhen Su\textsuperscript{1}, Yifan Li\textsuperscript{2}, Chao Jiang\textsuperscript{2} \\
$^1$School of Computer Engineering, Jiangsu Ocean University\\ $^2$School of Computer Science and Technology, Soochow University\\
}

\begin{document}
\maketitle
\begin{abstract}
To mitigate hallucinations in large language models (LLMs), we propose a framework that focuses on errors induced by prompts. Our method extends a chain-style knowledge distillation approach by incorporating a programmable module that guides knowledge graph exploration. This module is embedded as executable code within the reasoning prompt, allowing the model to leverage external structured knowledge during inference. Based on this design, we develop an enhanced distillation-based reasoning framework that explicitly regulates intermediate reasoning steps, resulting in more reliable predictions. We evaluate the proposed approach on multiple public benchmarks using GPT-4 and LLaMA-3.3.
Experimental results show that code-guided reasoning significantly improves contextual modeling and reduces prompt-induced hallucinations. Specifically, HIT@1, HIT@3, and HIT@5 increase by 15.64\%, 13.38\%, and 13.28\%, respectively, with scores exceeding 95\% across several evaluation settings. These findings indicate that the proposed method effectively constrains erroneous reasoning while improving both accuracy and interpretability.

\end{abstract}

\section{Introduction}

Large language models (LLMs) learn contextual representations from large-scale corpora, enabling them to capture rich semantic patterns and to perform a wide range of language-centric tasks, including natural language understanding, text generation, machine translation, and question answering \cite{brown2020language,raffel2020exploring}. Beyond text-only settings, recent advances have extended LLM capabilities to multimodal scenarios, allowing models to generate, translate, and reason over content spanning speech, images, and videos. This multimodal versatility has further accelerated their adoption across modern natural language processing systems and real-world applications \cite{bommasani2021opportunities}. In addition, emerging evidence suggests that LLMs can support reasoning-oriented tasks that require structured decision-making, multi-step inference, and contextual integration, rather than relying solely on shallow pattern matching or surface-level correlations \cite{yang2024harnessing, yang2024can, xiong2024large}.

Despite these capabilities, LLMs rely on probabilistic token prediction learned from training data. Because real-world corpora inevitably contain noise, bias, and incomplete information, models may produce outputs that are fluent but factually incorrect or logically inconsistent \cite{maynez2020faithfulness}. This behavior is commonly referred to as hallucination. One particularly challenging form is prompt-induced hallucination, which arises when ambiguous or misleading prompts lead to incorrect responses even for well-defined tasks \cite{ji2023survey, tonmoy2024comprehensive}.

\begin{figure*}[t]
  \centering
  \includegraphics[width=0.85\linewidth]{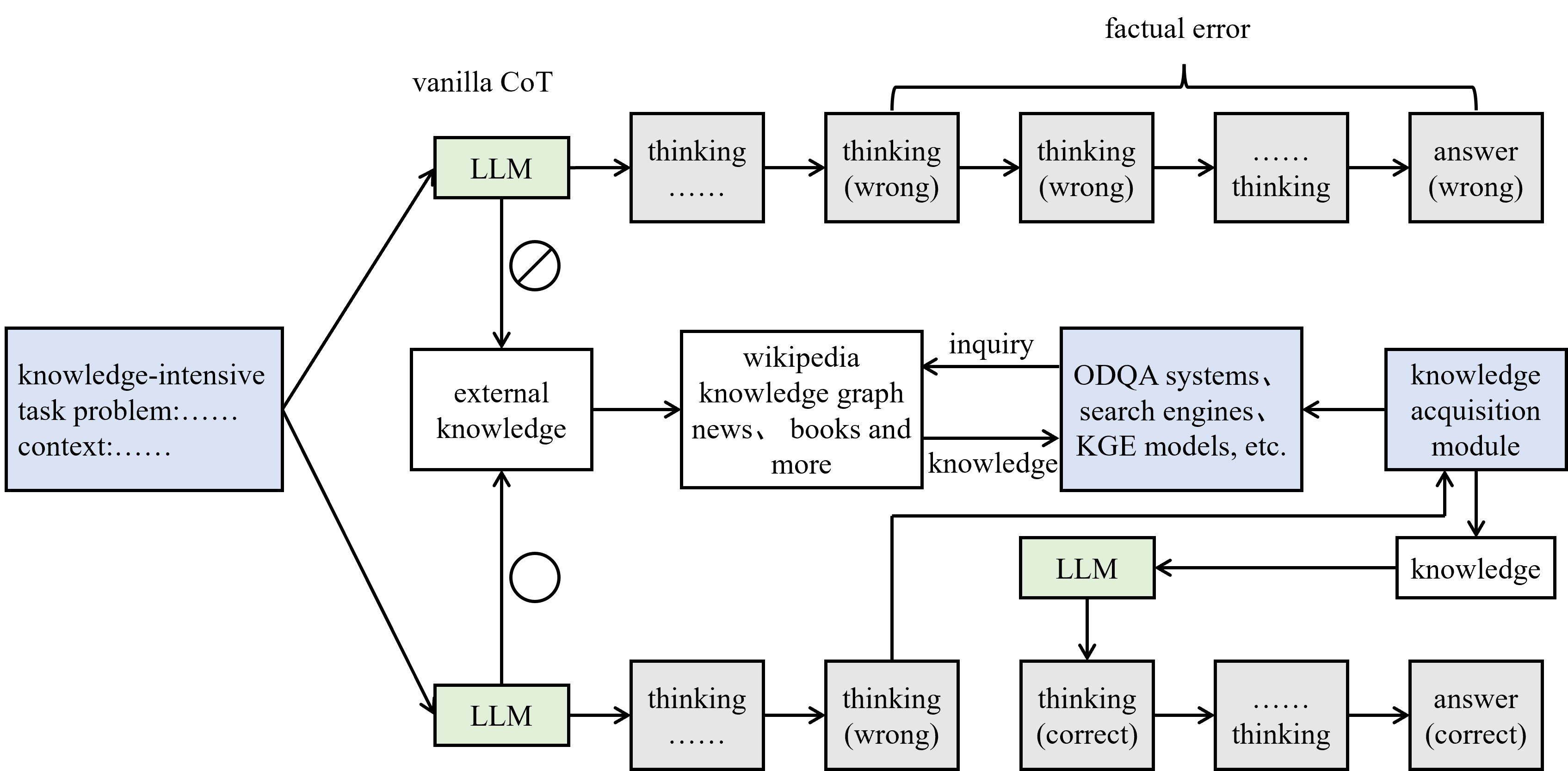}
  \caption{Structure of the knowledge distillation chain model.}

\end{figure*}

Hallucinations pose a major challenge to the reliable deployment of large language models (LLMs), especially in safety-critical domains such as scientific research, clinical decision support, and legal reasoning \cite{bender2021dangers}. Existing mitigation strategies include improving training data quality, augmenting generation with retrieved evidence, and applying post-hoc verification methods \cite{lewis2020retrieval,manakul2023selfcheckgpt}. While these approaches can reduce erroneous outputs, they often introduce additional computational cost or rely on external components, limiting their scalability and generality \cite{shuster2021retrieval}. Moreover, retrieval or verification alone does not explicitly constrain the model’s internal reasoning process, which is increasingly recognized as essential for reliable multi-step inference \cite{xiong2025deliberate}.

To address these limitations, we propose a prompt-induced hallucination mitigation method based on an enhanced knowledge distillation chain framework. By integrating structured knowledge with code-guided reasoning during inference, the proposed approach improves reasoning robustness while preserving the flexibility of large language models.

\section{Knowledge Distillation Chain-Style Model}

Knowledge distillation chain-style approaches integrate distillation principles with chain-of-thought reasoning to enhance both interpretability and predictive accuracy \cite{hinton2015distilling,wei2022chain}. By decomposing complex problems into a sequence of intermediate reasoning steps, this paradigm allows large language models to produce more structured and logically coherent outputs \cite{kojima2022large}.

In a typical distillation chain-based setting, a model processes an input query by generating intermediate reasoning traces prior to emitting a final answer. These traces act as explicit guidance signals that steer the model toward more reliable conclusions \cite{wang2023selfconsistency}. However, when intermediate reasoning depends exclusively on the model’s internal knowledge, errors can accumulate across steps, ultimately resulting in hallucinated outcomes \cite{ji2023survey}. 

To mitigate this issue, we augment the distillation chain framework with external structured knowledge sources, such as knowledge graphs that explicitly encode entities, relations, and temporal constraints \cite{xiongtilp,xiong2024teilp}. The reasoning process is enriched with auxiliary constraints that regulate intermediate steps and reduce reliance on uncertain internal representations, thereby improving logical consistency across multi-step inference \cite{nye2021show}.

\section{Improved Knowledge Distillation Chain with Code Guidance}

\subsection{Model Enhancement}

The enhanced knowledge distillation chain-based model incorporates a programmable component into the reasoning workflow. This component guides knowledge exploration by using code as an explicit control signal, allowing the model to regulate reasoning steps beyond natural language alone.

The code-driven component fulfills two key functions. First, it enables structured traversal of relevant knowledge, supporting systematic retrieval and reasoning over related entities and relations. Second, code is embedded within the reasoning prompts as an auxiliary representation, providing external structured information that complements textual reasoning.

Through code-guided reasoning, intermediate inference steps are more closely aligned with formal constraints and structured knowledge, which helps suppress hallucinated reasoning trajectories.

\begin{figure}[t]
  \centering
  \includegraphics[width=0.5\linewidth]{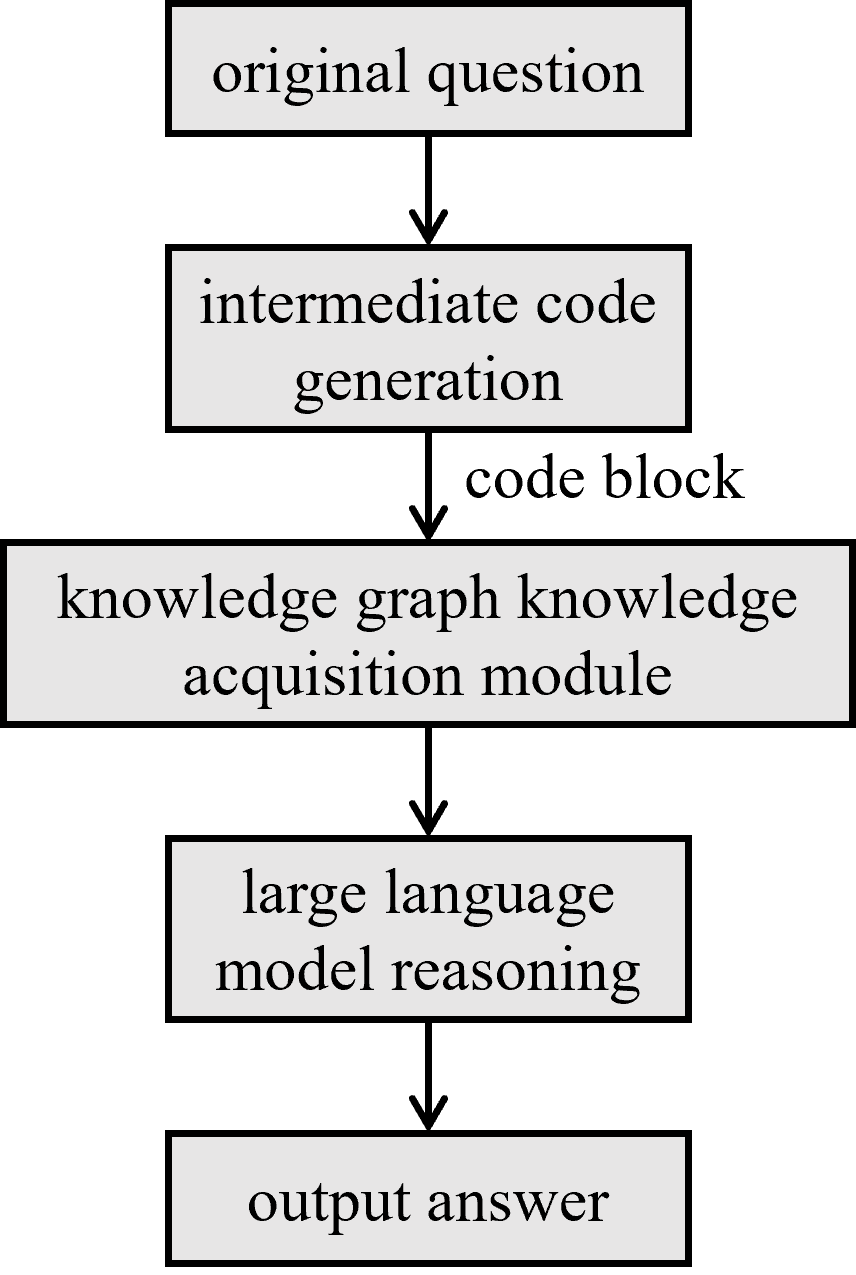}
  \caption{The process of suggesting hallucination problem-solving methods based on the large model based on the improved knowledge distillation chain.}
\end{figure}

\subsection{Reasoning Process Analysis}

With the enhanced knowledge distillation chain framework, we examine how large language models perform inference. The structured reasoning process enables validation of intermediate conclusions and supports the identification of inconsistencies during generation. As a result, the model exhibits stronger self-correction behavior and improved prediction accuracy.

In addition, the explicit organization of reasoning steps increases transparency and interpretability, facilitating clearer analysis of error sources in model outputs.

\subsection{Prompt-Induced Hallucination Mitigation Method}

Building on the enhanced knowledge distillation chain framework, we introduce a method for mitigating prompt-induced hallucinations in large language models. The approach combines structured reasoning with guidance from external knowledge to limit incorrect generation caused by underspecified or ambiguous prompts.

The mitigation procedure follows three steps. First, the input prompt is examined and reformulated into a set of structured sub-problems. Next, a code-guided distillation chain produces intermediate reasoning under external constraints. Finally, the model generates an answer grounded in validated reasoning steps and structured knowledge.

By limiting error accumulation across reasoning stages, the proposed method improves robustness to variations in prompt formulation.

\begin{figure}[t]
  \centering
  \includegraphics[width=\linewidth]{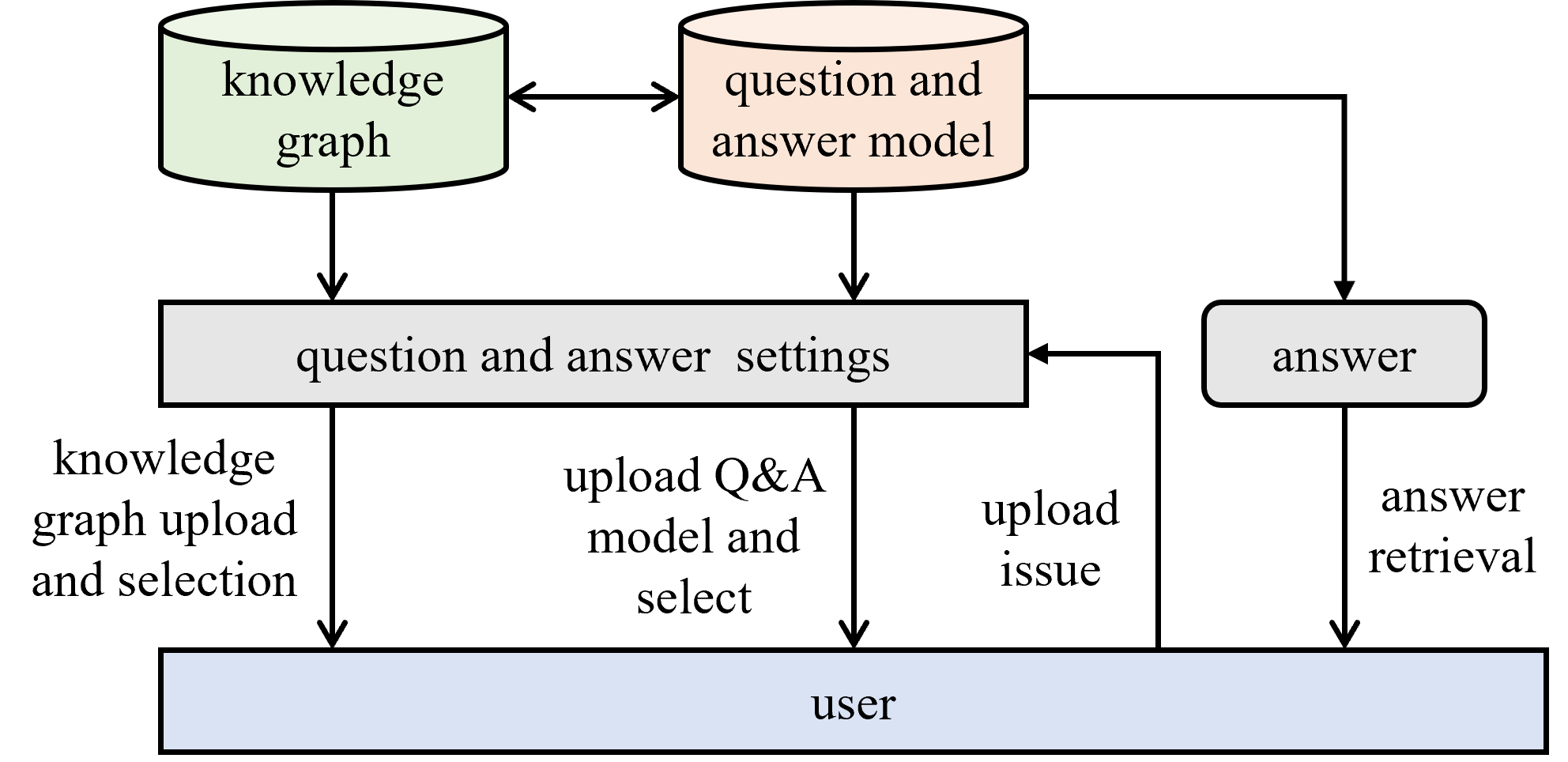}
  \caption{Simulation experiments.}
  \label{Fig5}
\end{figure}

\section{Experiments}

We evaluate the proposed approach on several public benchmarks using large language models such as GPT-4 and LLaMA 3.3 as base systems \cite{openai2023gpt4,touvron2023llama}. Model performance is assessed with standard retrieval-based metrics, including HIT@1, HIT@3, and HIT@5 \cite{bordes2013translating}.

The results show that incorporating code-guided reasoning leads to clear improvements in contextual understanding. Relative to baseline methods, the proposed framework yields consistent gains across all metrics, with HIT@1, HIT@3, and HIT@5 exceeding 95 percent. This indicates a substantial reduction in errors caused by prompt-induced hallucinations \cite{ji2023survey}.

Overall, the findings demonstrate that the enhanced knowledge distillation chain framework improves both prediction accuracy and output verifiability, aligning with prior evidence on the benefits of structured reasoning and external guidance \cite{nye2021show}.

\begin{table*}[t]
\centering
\caption{Improvement Verification Results of the Knowledge Distillation Chain Model (KDCM) (\%)}
\label{tab:improvement_verification}
\resizebox{0.7\linewidth}{!}{%
\begin{tabular}{l l c c c}
\toprule
\textbf{Dataset} & \textbf{Model} & \textbf{HIT@1} & \textbf{HIT@3} & \textbf{HIT@5} \\
\midrule
\multirow{2}{*}{WebQSP}
& KDCM
& 82.36 & 83.14 & 80.26 \\
& KDCM + Code Module
& 99.33 & 97.38 & 95.28 \\
\midrule
\multirow{2}{*}{CWQ}
& KDCM
& 81.36 & 82.09 & 82.14 \\
& KDCM + Code Module
& 97.86 & 98.03 & 96.20 \\
\midrule
\multirow{2}{*}{GSM8K}
& KDCM
& 82.06 & 85.79 & 84.39 \\
& KDCM + Code Module
& 98.23 & 95.14 & 95.47 \\
\midrule
\multirow{2}{*}{MWP}
& KDCM
& 85.26 & 84.39 & 82.11 \\
& KDCM + Code Module
& 98.19 & 96.78 & 95.08 \\
\midrule
\multirow{2}{*}{Dr.\ SPIDER}
& KDCM
& 86.29 & 83.14 & 85.29 \\
& KDCM + Code Module (Ours)
& 94.10 & 93.22 & 92.18 \\
\bottomrule
\end{tabular}
}
\end{table*}

\begin{figure*}[t]
  \centering
  \includegraphics[width=0.9\linewidth]{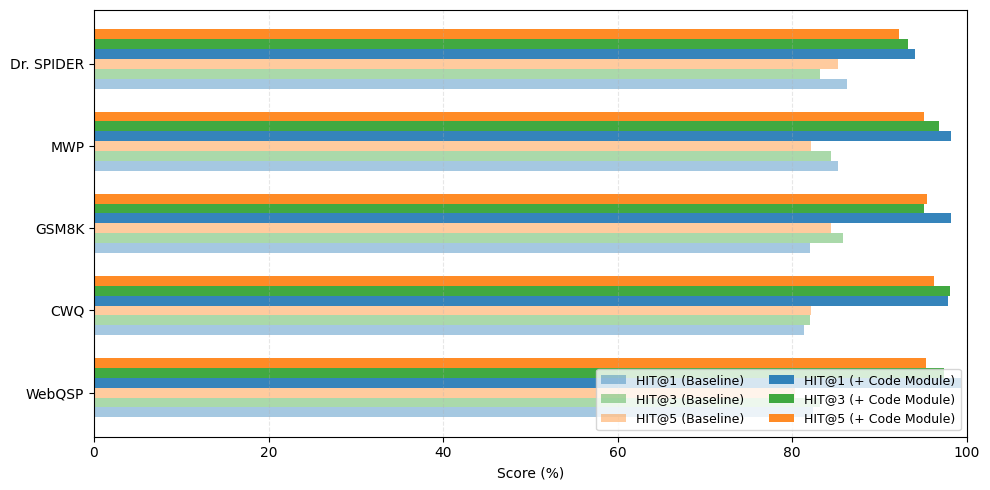}
  \caption{Verification results of the improvement of the knowledge distillation chain model.}
\end{figure*}

The experiments in this paper use publicly available datasets, including web-based question–answering datasets (WebQuestionsSP, WebQSP), CWQ (Complex Web Questions), GSM8K, MWP (Math Word Problems), and the Dr. SPIDER dataset, to evaluate the performance of the proposed method \citep{
yih2016value,
talmor2018web,
cobbe2021training,
koncel2015parsing,
li2023drspider,
berant2013semantic}.

\begin{table}[t]
\centering
\caption{Mean Evaluation Metrics of Different Methods on Experimental Datasets (\%)}
\label{tab:mean_evaluation_methods}
\resizebox{0.95\linewidth}{!}{%
\begin{tabular}{l c c c}
\toprule
\textbf{Method} & \textbf{HIT@1} & \textbf{HIT@3} & \textbf{HIT@5} \\
\midrule
Average (Ours) & 98.40 & 96.83 & 95.51 \\
KG-LLM-PR & 91.06 & 91.78 & 90.22 \\
LLM-SubKG-Sum & 92.23 & 91.89 & 90.17 \\
RAG & 90.23 & 90.28 & 90.18 \\
Self-Check & 91.25 & 92.35 & 91.27 \\
\bottomrule
\end{tabular}
}
\end{table}

\begin{table}[t]
\centering
\caption{Generalization Verification Results (\%)}
\label{tab:generalization_verification}
\resizebox{0.95\linewidth}{!}{%
\begin{tabular}{l c c c}
\toprule
\textbf{Method} & \textbf{HIT@1} & \textbf{HIT@3} & \textbf{HIT@5} \\
\midrule
Proposed Method & 99.18 & 97.64 & 95.12 \\
KG-LLM-PR & 90.26 & 88.52 & 86.47 \\
LLM-SubKG-Sum & 92.36 & 90.11 & 86.25 \\
RAG & 90.36 & 90.25 & 91.09 \\
Self-Check & 90.28 & 91.41 & 91.26 \\
\bottomrule
\end{tabular}
}
\end{table}

\subsection{Results and Analysis}
We assess the proposed approach using GPT-4 and LLaMA 3.3 as representative large language models \cite{openai2023gpt4,touvron2023llama}. For each benchmark, we compare the original model with its enhanced counterpart built on the improved knowledge distillation chain framework. Additional comparisons are conducted against KG-LLM-PR \citep{zhang2025kgllm} and LLM-SubKG-Sum \citep{zhang2024kgentity}. All systems are evaluated under identical inference conditions, and performance is measured using HIT@1, HIT@3, and HIT@5.

The experimental results show that the proposed method consistently outperforms baseline approaches across all datasets. The enhanced distillation chain framework yields clear gains on all evaluation metrics, reflecting a notable reduction in prompt-induced hallucinations.

Further analysis indicates that code-guided reasoning strengthens the model’s ability to capture and exploit contextual information. By regulating intermediate reasoning with structured knowledge, the model reduces dependence on uncertain internal representations and produces outputs that are both more accurate and easier to verify.

Across diverse datasets, the method demonstrates stable performance improvements, suggesting strong generalization. The advantages are especially evident on tasks involving multi-step reasoning, where hallucination errors tend to accumulate under standard chain-of-thought inference.

\subsection{Robustness Analysis}

To evaluate robustness, we test the proposed approach under different prompt formulations and dataset conditions. The results show that the enhanced model preserves strong HIT@K performance even when prompts are underspecified or ambiguous.

This robustness stems from the structured reasoning mechanism imposed by the improved knowledge distillation chain framework. By explicitly regulating intermediate inference steps, the model becomes less affected by prompt noise and limits the accumulation of reasoning errors.

\subsection{Generalization Evaluation}

We further assess the generalization capability of the proposed approach by testing it on benchmarks distinct from those used during model development. The results show consistent performance across domains, with high accuracy and reduced hallucination behavior.

In comparison with existing hallucination mitigation techniques, the proposed method delivers stronger results across multiple evaluation settings. These findings suggest that the approach does not depend on dataset-specific heuristics and can be broadly applied to diverse large language model tasks.

\section{Conclusion}

This work introduces a method for mitigating prompt-induced hallucinations in large language models through an enhanced knowledge distillation chain framework. By integrating code-guided reasoning with structured external knowledge during inference, the proposed approach improves reasoning reliability, robustness, and interpretability. Experimental results across multiple public benchmarks show consistent improvements, with HIT@1, HIT@3, and HIT@5 surpassing 95 percent in several evaluation scenarios, demonstrating a clear reduction in hallucination behavior while preserving model flexibility. Future research will explore extending this framework to multimodal reasoning settings and combining it with retrieval-augmented generation and reinforcement learning based optimization strategies.

\section*{Limitations}

Although the proposed approach is effective, it has several limitations. First, incorporating code-guided reasoning increases inference complexity and can introduce additional computational cost compared with standard prompting methods. Second, the approach depends on access to structured external knowledge and suitable code representations, which may restrict its use in domains where such resources are limited. Finally, while strong results are observed on text-based reasoning benchmarks, further validation is needed to assess performance in open-ended generation and multimodal settings.

\bibliography{custom}

\end{document}